# Multi-Sensor Perceptual System for Mobile Robot and Sensor Fusion-based Localization

T. T. Hoang, P. M. Duong, N. T. T. Van, D. A. Viet and T. Q. Vinh

Department of Electronics and Computer Engineering
University of Engineering and Technology
Vietnam National University, Hanoi

*Abstract -* **This paper presents an Extended Kalman Filter (EKF) approach to localize a mobile robot with two quadrature encoders, a compass sensor, a laser range finder (LRF) and an omni-directional camera. The prediction step is performed by employing the kinematic model of the robot as well as estimating the input noise covariance matrix as being proportional to the wheel's angular speed. At the correction step, the measurements from all sensors including incremental pulses of the encoders, line segments of the LRF, robot orientation of the compass and deflection angular of the omni-directional camera are fused. Experiments in an indoor structured environment were implemented and the good localization results prove the effectiveness and applicability of the algorithm.**

## I. INTRODUCTION

A mobile autonomous robot requires accurate positioning for its navigational action. From current readings of the sensors, robot must determine exactly its position and orientation in the environment. There are some kind of sensors which can be used for robot. Each sensor often only measure one or two environmental parameters with a limited accuracy. Naturally, more used sensor, more accuracy for determine position of robot. That is reason the sensor fusion method has been carried out in modern robots in order to increase the accuracy of measurements. Almost implementation of this method are based on probabilistic inferences. The Extended Kalman Filter (EKF) is the most efficient probabilistic solution to simultaneously estimate the robot position and orientation based on some interoceptive and exteroceptive sensor information.

Mobile robots often use optical encoders as interoceptive sensor for determining of position following a method named as the odometry. However, due to accumulated error, the uncertainty of estimated position by odometric system is increased time by time during robot's moving. In order to overcome this disadvantage, by using a Kalman filter with other measurements from one or some exteroceptive sensor combined with odometric measures, estimated position value becomes more accurate. That means the estimated trajectory is nearer to the true trajectory.

Several works have been reported to cope with the problem of sensor fusion. Ying Zhang has developed a Bayesian technique to estimate the state evolution, which prevents from dealing with Dirac delta function [1]. Al-Dhaher and Makesy proposed a generic architecture that employed an standard Kalman filter and fuzzy logic techniques to adaptively incorporate sensor measurements in a high accurate manner [2]. In [3], a Kalman filter update equation was developed to obtain the correspondence of a line segment to a model, and this correspondence was then used to correct position estimation. In [4], an extended Kalman filter was conducted to fuse the DGPS and odometry measurements for outdoor-robot navigation.

In our work, one novel platform of mobile robot with some modern sensor was designed and installed. A positioning software was developed which include the Extended Kalman Filter (EKF). The input information is an odometric system, a compass sensor, a LRF and an omni-directional camera. The output information is the robot's pose including the position and orientation. The experiment shows that the estimated output values is nearest to the true trajectories when measurements from all sensors are fused together.

The paper is arranged as follows. Details of the sensor system are described in Section II. The algorithm for sensor fusion using EKF is explained in Section III. Section IV introduces experimental results. The paper concludes with an evaluation of the system.

## II. SENSOR SYSTEM DESIGN

The sensor system consists of a compass sensor, a LRF, an Omni-directional camera and three quadrature encoders. Fig.1 describes the sensors in relation with communication channels in a mobile robot.

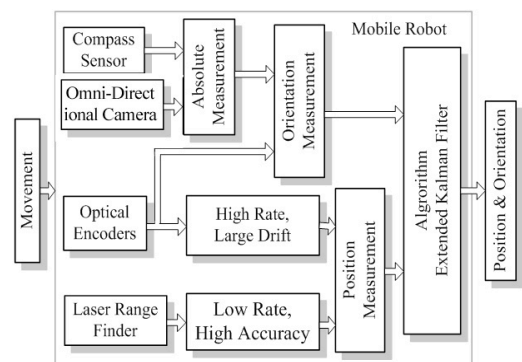

Figure 1. Configuration of the sensor system

The *optical quadrature encoders* are used as measurement for positioning and feedback for a closed-loop motor speed controller. An optical encoder is basically a mechanical light chopper that produces a certain number of sine or square wave pulses for each shaft revolution. As the diameter of wheel and the gear coefficient are known, the angular position and speed of wheel can be calculated. In the quadrature encoder, a second light chopper is placed 90 degrees shifted with respect to the original resulting in twin square waves as shown in fig.2. Observed phase relationship between waves is employed to determine the direction of the rotation.

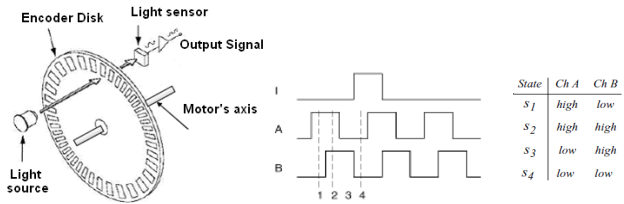

Figure 2. Optical encoder structure and output pulses

The heading sensor is used to determine the robot orientation. This sensory module contains a CMPS03 compass sensor operating based on Hall effect with heading resolution of 0.1° (fig.3). The module has two axes, x and y. Each axis reports the strength of the magnetic field component paralleled to it. The microcontroller connected to the module uses synchronous serial communication to get axis measurements [5].

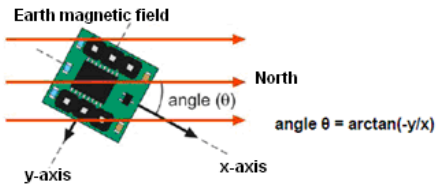

Figure 3. Heading module and output data

The omni-directional camera is a Hyper-Omni Vision SOIOS 55. It consists of a hyperbolical mirror placed a short distance above a normal camera to acquire a 360-degree view around the robot (fig.4).

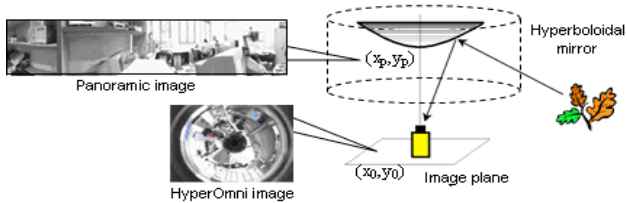

Figure 4. Operation of the omni-directional camera and its captured images

From the center of the omni-directional image $(x_c, y_c)$, the coordinate relation between a pixel in the omni-directional image $(x_0, y_0)$ and its projection in the cylinder plane $(x_p, y_p)$ is as follows:

$$\begin{matrix} x_0 = x_c + r\sin\theta \\ y_0 = x_c + r\cos\theta \end{matrix} \text{ where } r = y_p \text{ and } \theta = \frac{x_p.360}{720} = \frac{x_p}{2} \quad (1)$$

Images captured by the omni-directional camera contains rich information about the position and orientation of objects. The main advantages of the camera include the continuous appearance of landmarks in the field view and the conservation of line feature over image transformations [8]. The quality of images however much depends on the lighting condition and may appear random noises. In addition, the resolution of the image is not uniform due to nonlinear characteristic of the hyperbolical mirror. Consequently, care should be maintain when performing image processing algorithms.

In this work, the omni-directional camera is used as an absolute orientation measurement and is fused with other sensor to create a complete perceptual system for the robot. The method is based on the detection of a red vertical landmarks located at a fixed position $(x_m, y_m)$. The conservation of line feature ensures that the shape of the landmark is unchanged in both omni-directional and panorama images. The robot's orientation determination then becomes the problem of calculating the orientation $x_p/2$ of the landmark (fig.5). The algorithm can be summarized as follows: From the capture image, a digital filter is applied to eliminate random noises. The red area is then detected and the image is transformed from the RGB color space to the grey scale. Applying Hough transform, the vertical line is extracted and the value $x_p/2$ corresponding to the robot's orientation is obtained.

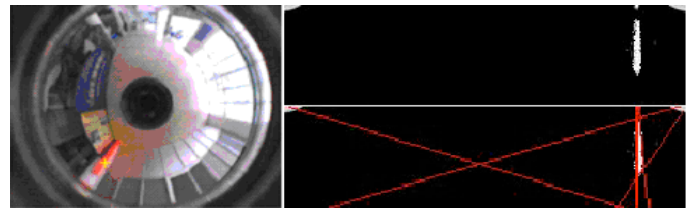

Figure 5. Line detection using Houh transform

A 2D laser range finder (LRF) with a range from 0.04m to 80m is developed for the system. Its operation is based on the time-of-flight measurement principle. A single laser pulse is emitted out and reflected by an object surface within the range of the sensor. The lapsed time between emission and reception of the laser pulse is used to calculate the distance between the object and the sensor. By an integrated rotating mirror, the laser pulses sweep a radial range in its front so that a 2D field/area is defined [10]. In each environment scan, the LRF gives a set of distance data to the obstacles at the angles. Fig.6 shows the view field of one LRF scan. Based on this data, environmental features can be found [9].

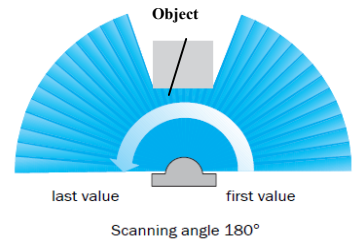

Figure 6. The view field of a LRF scan

In each environment scan, the LRF gives a set of distances $d_i=[d_0, ..., d_{180}]$ to the obstacles at the angles $\beta_i = [0^0, ..., 180^0]$. The line segments are then extracted from the reflection points. Fig.7 outlines the problem of visibility.

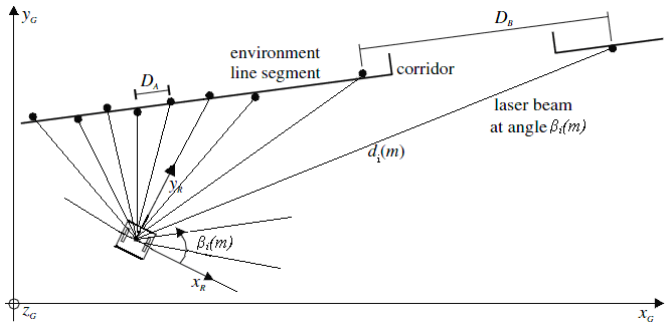

Figure 7. Distance data collected at each scan of LRF

Fig.8 shows the proposed sensor system implemented in a mobile robot developed at our laboratory.

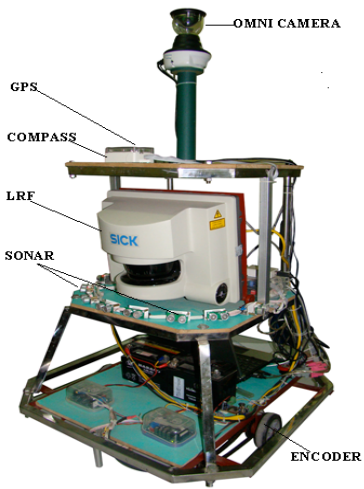

Figure 8. The implemented sensor system

### III. SENSOR FUSION

The proposed sensor platform equips the robot with the ability to perceive many parameters of the environment. Their combination, however, presents even more useful information. In this work, the raw data of three different types of sensors including the compass sensor, the quadrature encoder and the LRF is syndicated inside an EKF. The aim is to determine the robot position during operation as accurately as possible.

We start with the kinematic model of the mobile robot. The two wheeled, differential-drive mobile robot with non-slipping and pure rolling is considered. Fig.9 shows the coordinate systems and notations for the robot, where $(X_G, Y_G)$ is the global coordinate, $(X_R, Y_R)$ is the local coordinate relative to the robot chassis. $R$ denotes the radius of driven wheels, and $L$ denotes the distance between the wheels.

During one sampling period $\Delta t$, the rotational speed of the left and right wheels $\omega_L$ and $\omega_R$ create corresponding increment distances $\Delta s_L$ and $\Delta s_R$ traveled by the left and right wheels respectively:

$$\Delta s_L = \Delta t R \omega_L \qquad \Delta s_R = \Delta t R \omega_R \qquad (1)$$

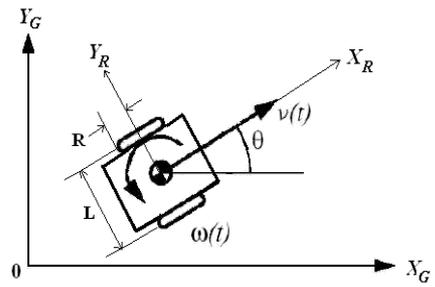

Figure 9. The robot's pose and parameters

These can be translated to the linear incremental displacement of the robot's center $\Delta s$ and the robot's orientation angle $\Delta \theta$:

$$\Delta s = \frac{\Delta s_L + \Delta s_R}{2}$$
$$\Delta \theta = \frac{\Delta s_R - \Delta s_L}{L} \qquad (2)$$

The coordinates of the robot at time $k+1$ in the global coordinate frame can be then updated by:

$$\begin{bmatrix} x_{k+1} \\ y_{k+1} \\ \theta_{k+1} \end{bmatrix} = \begin{bmatrix} x_k \\ y_k \\ \theta_k \end{bmatrix} + \begin{bmatrix} \Delta s_k \cos(\theta_k + \Delta \theta_k / 2) \\ \Delta s_k \sin(\theta_k + \Delta \theta_k / 2) \\ \Delta \theta_k \end{bmatrix} \qquad (3)$$

In practice, (3) is not really accurate due to unavoidable errors appeared in the system. Errors can be both systematic such as the imperfectness of robot model and nonsystematic such as the slip of wheels. These errors have accumulative characteristic so that they can break the stability of the system if appropriate compensation is not considered. In our system, the compensation is carried out by the EKF.

Let $\mathbf{x} = [x\,y\,\theta]^T$ be the state vector. This state can be observed by some absolute measurements, $\mathbf{z}$. These measurements are described by a nonlinear function, $h$, of the robot coordinates and an independent Gaussian noise process, $\mathbf{v}$. Denoting the function (3) is $f$, with an input vector $\mathbf{u}$, the robot can be described by:

$$\mathbf{x}_{k+1} = f(\mathbf{x}_k, \mathbf{u}_k, \mathbf{w}_k)$$
$$\mathbf{z}_k = h(\mathbf{x}_k, \mathbf{v}_k) \qquad (4)$$

where the random variables $\mathbf{w}_k$ and $\mathbf{v}_k$ represent the process and measurement noise respectively. They are assumed to be independent to each other, white, and with normal probability distributions: $\mathbf{w}_k \sim \mathbf{N}(0, \mathbf{Q}_k) \quad \mathbf{v}_k \sim \mathbf{N}(0, \mathbf{R}_k) \quad E(\mathbf{w}_i \mathbf{v}_j^T) = 0$

The steps to calculate the EKF are then realized as below:

1. Prediction step with time update equations:

$$\hat{\mathbf{x}}_k^- = f(\hat{\mathbf{x}}_{k-1}, \mathbf{u}_{k-1}, \mathbf{0}) \qquad (5)$$
$$\mathbf{P}_k^- = \mathbf{A}_k \mathbf{P}_{k-1} \mathbf{A}_k^T + \mathbf{W}_k \mathbf{Q}_{k-1} \mathbf{W}_k^T \qquad (6)$$

where $\hat{\mathbf{x}}_k^- \in \Re^n$ is the *priori* state estimate at step $k$ given knowledge of the process prior to step $k$, $\hat{\mathbf{P}}_k^-$ denotes the covariance matrix of the state-prediction error, $\mathbf{A}_k$ is the Jacobian matrix of partial derivates of $f$ to $x$:

$$\mathbf{A}_{ij} = \frac{\partial \mathbf{f}_i}{\partial \hat{\mathbf{x}}_{pj(k-1)}}\bigg|_{(\hat{\mathbf{x}}_{p(k-1)}, \mathbf{u}_{(k-1)})}; \mathbf{A}_k = \begin{bmatrix} 1 & 0 & -\Delta s_k \sin(\theta_k + \Delta\theta_k/2) \\ 0 & 1 & \Delta s_k \cos(\theta_k + \Delta\theta_k/2) \\ 0 & 0 & 1 \end{bmatrix} \quad (7)$$

**W** is the Jacobian matrix of partial derivates of $f$ to $w$:

$$\mathbf{W}_{ij} = \frac{\partial \mathbf{f}_i}{\partial \mathbf{w}_{j(k-1)}}\bigg|_{(\hat{\mathbf{x}}_{p(k-1)}, \mathbf{u}_{(k-1)})}; \mathbf{W}_k = \begin{bmatrix} -\Delta S_k \sin(\theta_k + \Delta\theta_k/2)/2l & \Delta S_k \sin(\theta_k + \Delta\theta_k/2)/2l \\ \Delta S_k \cos(\theta_k + \Delta\theta_k/2)/2l & -\Delta S_k \cos(\theta_k + \Delta\theta_k/2)/2l \\ 1/l & -1/l \end{bmatrix} \quad (8)$$

and $\mathbf{Q}_{k-1}$ is the input-noise covariance matrix which depends on the standard deviations of noise of the right-wheel rotational speed and the left-wheel rotational speed. They are modeled as being proportional to the rotational speed $\omega_{R,k}$ and $\omega_{L,k}$ of these wheels at step $k$. This results in the variances equal to $\delta\omega_R^2$ and $\delta\omega_L^2$, where $\delta$ is a constant determined by experiments. The input-noise covariance matrix $\mathbf{Q}_k$ is defined as:

$$\mathbf{Q}_k = \begin{bmatrix} \delta\omega_{R,k}^2 & 0 \\ 0 & \delta\omega_{L,k}^2 \end{bmatrix} \quad (9)$$

2. Correction step with measurement update equations:

$$\mathbf{K}_k = \mathbf{P}_k^- \mathbf{H}_k^T (\mathbf{H}_k \mathbf{P}_k^- \mathbf{H}_k^T + \mathbf{R}_k)^{-1} \quad (10)$$

$$\hat{\mathbf{x}}_k = \hat{\mathbf{x}}_k^- + \mathbf{K}_k \left( \mathbf{z}_k - \mathbf{h}(\hat{\mathbf{x}}_k^-) \right) \quad (11)$$

$$\mathbf{P}_k = (\mathbf{I} - \mathbf{K}_k \mathbf{H}_k) \mathbf{P}_k^- \quad (12)$$

where $\hat{\mathbf{x}}_k \in \Re^n$ is the *posteriori* state estimate at step $k$ given measurement $\mathbf{z}_k$, $\mathbf{K}_k$ is the Kalman gain, **H** is the Jacobian matrix of partial derivates of $h$ to $x$:

$$\mathbf{H}_{ij} = \frac{\partial \mathbf{u}_i}{\partial \hat{\mathbf{x}}_{pj(k-1)}}\bigg|_{(\hat{\mathbf{x}}_{p(k)})}; \mathbf{H}_k = \begin{bmatrix} -C_1 \cos(\beta_1) & -C_1 \sin(\beta_1) & 0 \\ 0 & 0 & -1 \\ . & . & . \\ . & . & . \\ . & . & . \\ -C_N \cos(\beta_N) & -C_N \sin(\beta_N) & 0 \\ 0 & 0 & -1 \\ 0 & 0 & 1 \\ 0 & 0 & 1 \end{bmatrix} \quad (13)$$

$\mathbf{R}_k$ is the covariance matrix of noises estimated from the measurements of compass sensor and LRF as follow.

At the first scan of LRF, a global map of environment is constructed composed of a set of line segments described by parameters $\beta_j$ and $\rho_j$. The line equation in normal form is:

$$x_G \cos\beta_j + y_G \sin\beta_j = \rho_j \quad (14)$$

When the robot moves, a new scan of LRF is performed and a new map of environment, namely local map, is constructed which also consists of a set of line segments described by the equation:

$$x_R \cos\psi_i + y_R \sin\psi_i = r_i \quad (15)$$

where $\psi_i$ and $r_i$ are the parameters of lines (fig.10).

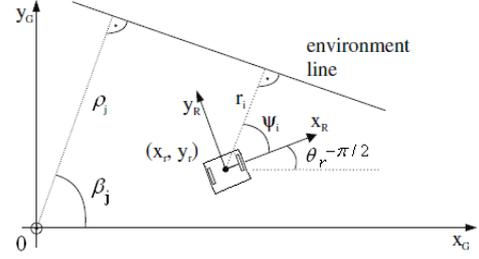

Figure 10. The line parameters ($\rho_j$, $\beta_j$) according to the global coordinates, and the line parameters ($r_i$, $\psi_i$) according to the robot coordinates

The line segments of the global and local map are then matched using *weighted line fitting algorithm* [11]. The matching line parameters $r_i$ and $\psi_i$ from the local map are collected in the vector $\mathbf{z}_k$, which is used as the input for the correction step of the EKF to update the robot's state:

$$\mathbf{z}_k = [r_1, \psi_1, \ldots, r_N, \psi_N, \varphi_k, \gamma_k]^T \quad (16)$$

where $\varphi_k$ and $\gamma_k$ are orientation measurements of the compass sensor and the omni-directional camera, respectively.

From the robot pose estimated by odometry, the parameters $\rho_j$ and $\beta_j$ of the line segments in the global map (according to the global coordinates) is transformed into the parameters $\hat{r}_i$ and $\hat{\psi}_i$ (according to the coordinates of the robot) by:

$$C_j = \rho_j - x_{r(k)} \cos(\beta_j) - y_{r(k)} \sin(\beta_j) \quad (17)$$

$$\begin{bmatrix} \hat{r}_i \\ \hat{\psi}_i \\ \hat{\theta}_k \\ \hat{\theta}_k \end{bmatrix} = \mathbf{u}(\hat{x}_k^-, \rho_j, \beta_j, \varphi_i, \gamma_i) = \begin{bmatrix} |C_j| \\ \beta_j - (\hat{\theta}_i^- - \pi/2) + (-0.5\,\text{sign}(C_j) \quad 0.5)\pi \\ \hat{\theta}_k^- \\ \hat{\theta}_k^- \end{bmatrix} \quad (18)$$

The covariance matrix $\mathbf{R}_k$ of measurement noise has the diagonal structure, The noise of wheel speed measures can be determined by experiment. The accuracy of compass sensor and LRF measures can be received from the manufacture technical specification. These are filled in $R_k$ for the correction step of EKF.

$$\mathbf{R}_k \cong \begin{bmatrix} var(r) & 0 & \ldots & 0 & 0 & 0 \\ \ldots & var(\psi) & \ldots & \ldots & \ldots & \ldots \\ \ldots & \ldots & \ldots & \ldots & \ldots & \ldots \\ \ldots & \ldots & \ldots & var(\psi) & \ldots & \ldots \\ \ldots & \ldots & \ldots & \ldots & var(\varphi) & 0 \\ 0 & 0 & 0 & \ldots & 0 & var(\gamma) \end{bmatrix} \quad (19)$$

From (16) and (19), the data of the compass sensor is utilized to correct the robot orientation. At step $k$, this data is employed as the absolute measurement for the element $θ_k$ of $z_k$. The noise of this measurement is achieved from the manufactory specification. The accuracy of $0.1^0$ corresponds to the noise variance of $0.01$. This collected into the covariance matrix $R_{i,k}$ (19) gives $R_k$ for the correction step of EKF.

## IV. EXPERIMENTS

The setup of the sensor system has been described in Section II. In this section, experiments are conducted to evaluate the efficiency of the fusion algorithm.

### A. Experimental Setup

A rectangular shaped flat-wall environment constructed from several wooden plates surrounded by a cement wall is setup. The robot is a two wheeled, differential-drive mobile robot. Its wheel diameter is *10 cm* and the distance between two drive wheels is *60cm*.

The speed stability of the motor during each sampling time step is an important factor to maintain the efficiency of the EKF. For that reason, motors are controlled by microprocessor-based electronic circuits which carries out a PID algorithm. The stability checked by a measuring program written in LABVIEW is *±5%*.

The compass sensor has the accuracy of $0.1^0$. The LRF has the accuracy of *30mm* in distance and $0.25^0$ in deflect angle. The sampling time $Δt$ of the EKF is *100ms*. The proportional factor $δ$ of the input-noise covariance matrix $Q_k$ is experimentally estimated as follows. The deviations between the true robot position and the position estimated by the kinematic model when driving the robot straight forwards several times (from the minimum to the maximum tangential speed of the robot) is determined. The deviations between the true robot orientation and the orientation estimated by the kinematic model when rotating the robot around its own axis several times (from the minimum to the maximum angular speed of the robot) is also determined. Based on the error in position and orientation, the parameter $δ$ is calculated. In our system, the $δ$ is estimated to be the value 0.01. Fig.11 shows the robot and environment structure in experiments.

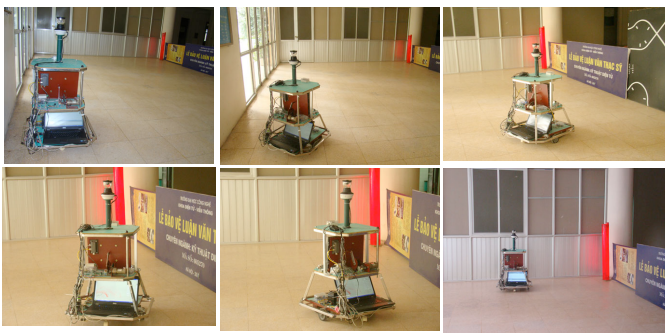

Figure 11. A sequence of images showing the motion of robot in an experimental environment during the autonomous navigation operation

### B. Sensor Fusion Algorithm Evaluation

In order to evaluate the efficiency of the fusion algorithm, different configurations of the EKF were implemented. Fig.12 describes the trajectories of a robot movement estimated by five methods: the odometry, the EKF with compass sensor, the EKF with LRF, the EKF with omni-directional camera and the EKF with the combination of all sensor. The deviations between each trajectory and the real one are represented in fig.13.

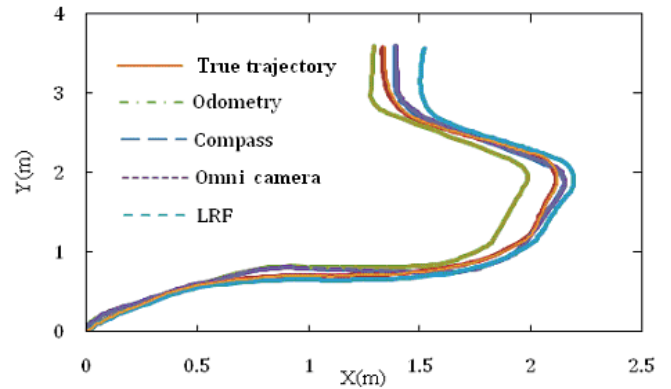

Figure 12. Estimated robot trajectories under different EKF configurations

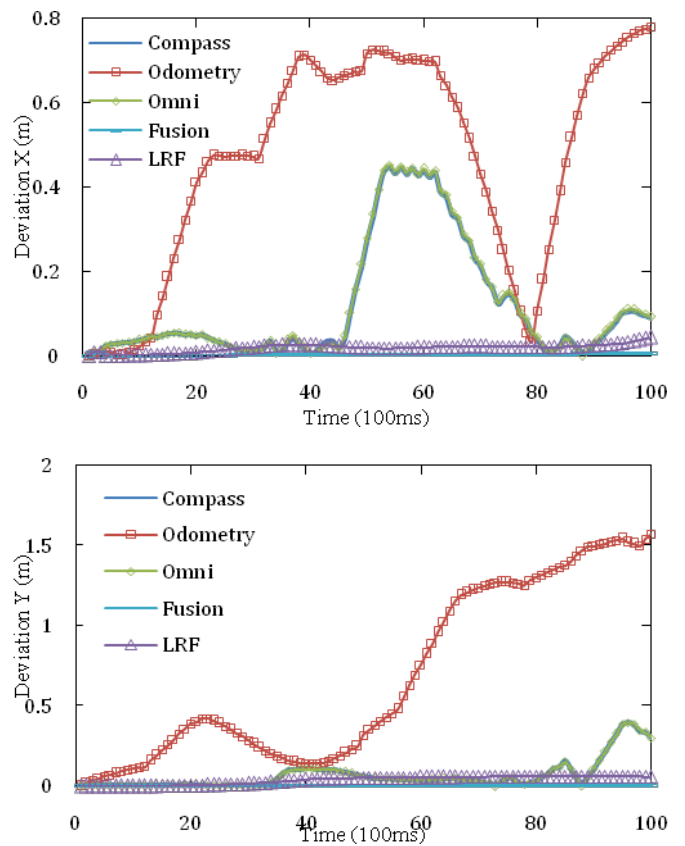

Figure 13. The deviation in X and Y direction between estimated positions and the real one

In another experiment, the robot is programmed to follow predefined paths under two different scenarios: with and without the EKF. Fig.14a represents the trajectories of the robot moving along a rounded rectangular path in which the one with dots corresponds to the non-existence of EKF and the one with pluses corresponds to the existence of EKF in the implementation. The trajectories in case the robot follows a rounded triangular path is shown in fig.14b.

It is concluded that the EKF algorithm improves the robot localization and the combination of all available sensors gives the optimal result.

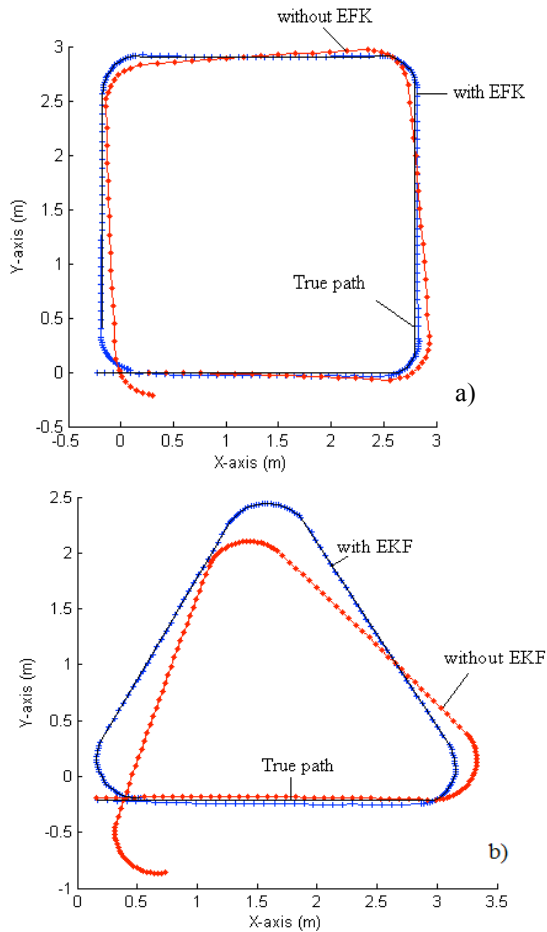

Figure 14. Trajectories of the robot moving along predefined paths with and without EKF [16]
a) Rounded rectangular path   b) Rounded triangular path

## V. CONCLUSION

It is necessary to develop a perceptual system for the mobile robot. The system is required to not only be well-working but also critically support various levels of perception. In this work, many types of sensors including position speed encoders, integrated sonar ranging sensors, compass and laser finder sensors, the global positioning system (GPS) and the visual system have been implemented in a real mobile robot platform. An EKF has been designed to fuse the raw data of compass sensor and LRF. Experiments show that this novel combination significantly improves the accuracy of robot localization and is sufficient to ensure the success of robot navigation. Further investigation will be continued with more sensor combination to better support the localization in outdoor environments.